\documentclass[a4paper]{svproc}
\usepackage{url}
\usepackage{graphicx}
\usepackage{array}
\usepackage{makecell}

\begin{document}
\mainmatter              % start of a contribution
\title{ROSS: Radar Off-road Semantic Segmentation}
\titlerunning{ROSS}  % abbreviated title (for running head)
%                                     also used for the TOC unless
%                                     \toctitle is used
%
\author{Peng Jiang\inst{1} \and Srikanth Saripalli\inst{2}}
\authorrunning{Peng Jiang et al.} % abbreviated author list (for running head)
%
%%%% list of authors for the TOC (use if author list has to be modified)
\tocauthor{Peng Jiang, Srikanth Saripalli}
\institute{Texas A\&M University, College Station TX 77843, USA,\\
\email{maskjp@tamu.edu},
\and
\email{ssaripalli@tamu.edu},
}

\maketitle              % typeset the title of the contribution

\begin{abstract}
As the demand for autonomous navigation in off-road environments increases, the need for effective solutions to understand these surroundings becomes essential. In this study, we confront the inherent complexities of semantic segmentation in RADAR data for off-road scenarios. We present a novel pipeline that utilizes LIDAR data and an existing annotated off-road LIDAR dataset for generating RADAR labels, in which the RADAR data are represented as images. Validated with real-world datasets, our pragmatic approach underscores the potential of RADAR technology for navigation applications in off-road environments.
\end{abstract}

\section{Introduction}
Robots' effective navigation and task performance in off-road environments is highly dependent on their understanding of the objects and structures around them. This ability requires the development of robust algorithms for semantic segmentation, which assign labels to each pixel or point in data obtained from images and LIDAR devices. The resultant data can facilitate tasks such as object recognition and scene comprehension, which is beneficial for a myriad of applications, such as inspection, exploration, rescue, and reconnaissance missions. 

In addition, the increasing demand for autonomous vehicle technology capable of optimally functioning in adverse weather conditions, such as rain and snow, underscores the need for effective solutions. While CAMERA and LIDAR sensors have shown promising performance under normal conditions, their efficiency plummets when faced with harsh weather. RADAR sensors, for example, those produced by Navtech \cite{barnes_oxford_2020}, present a potentially viable alternative, given their longer wavelengths and inherent resistance to small particles, such as fog, rain, or snow, that can affect the performance of CAMERA and LIDAR sensors.

Although RADAR sensors offer robust performance in inclement weather, they are not without shortcomings. Their lower spatial resolution and higher noise, compared to LIDAR sensors, exacerbate the difficulty of processing RADAR data. In contrast to other sensing modalities, such as CAMERA and LIDAR, RADAR data predominantly contain two-dimensional information, which introduces a significant challenge for semantic segmentation. 
\begin{figure}[t]
\centering
\includegraphics[width=\textwidth]{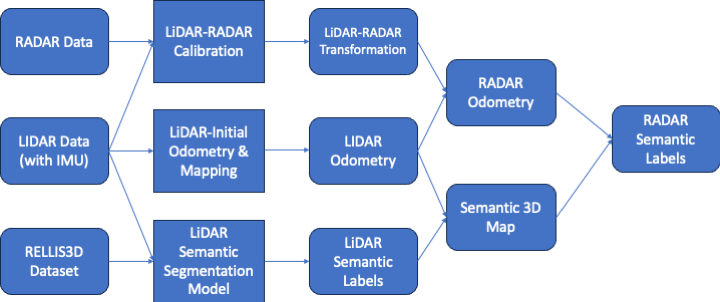}
\caption{Radar Label Generation Flow: 1) Calibration of the extrinsic transformation parameters between LIDAR and RADAR 2) Data collection from LIDAR, RADAR, and IMU sensors facilitated by the deployment of robots in off-road terrains
3) Application of semantic models to LIDAR data
4) Accumulation of LIDAR scans through a specialized LIDAR odometry algorithm
5) Fusion of LIDAR scan labels based on the resultant LIDAR odometry
6) Projection of LIDAR's semantic information onto RADAR using the calibrated extrinsic transformation parameters
7) Training of a model using the newly generated LIDAR labels} 
\label{fig:flow_label}
\end{figure}
The state-of-the-art semantic segmentation methods predominantly require manual labeling of ground truth data, a near-impossible feat with raw RADAR data. This reality has led to RADAR data receiving scant attention in the realm of semantic segmentation research. Additionally, the complexity of off-road environments makes it extremely challenging to manually label and interpret RADAR data, thereby thwarting attempts to train models using traditional supervised learning methods.
Initial attempts in RADAR semantic segmentation leveraged occupancy grid methods to discern between free and occupied spaces but failed to provide substantial information on object types \cite{weston_probably_2019}. To mitigate this, Lombacher et al. \cite{lombacher_semantic_2017} used a Fully Convolutional Neural Network (FCNN) to segment RADAR scans based on occupancy probability, using hand labels to classify objects. Scheiner et al. \cite{scheiner_radar-based_2018} extracted 50 distinct features from low-level RADAR cube data and measurement data to perform multiclass classification using random forest and long-short-term memory (LSTM) classifiers. Based on this, Scheiner et al. \cite{scheiner_radar-based_2019} increased the number of features to 98, implementing recurrent neural networks (RNNs) and classifier ensembles to improve classification accuracy. Unlike the use of hand-made features, Schumann et al. \cite{schumann_semantic_2018} regarded RADAR point clouds as regular point clouds and adapted PointNet++ to perform semantic segmentation directly. To counter the laborious process of manual labeling, Kaul et al. \cite{kaul_rss-net_2020} used weak supervision to label RADAR data by semantically segmenting camera streams and collating the results with LIDAR range measurements to generate labeled images. However, the majority of these works predominantly concentrate on urban scenes, with off-road environments being largely unexplored due to their inherently unstructured and less interpretable RADAR data.
\begin{figure}[t]
\centering
\begin{tabular}{c c} 
\includegraphics[width=0.5\textwidth]{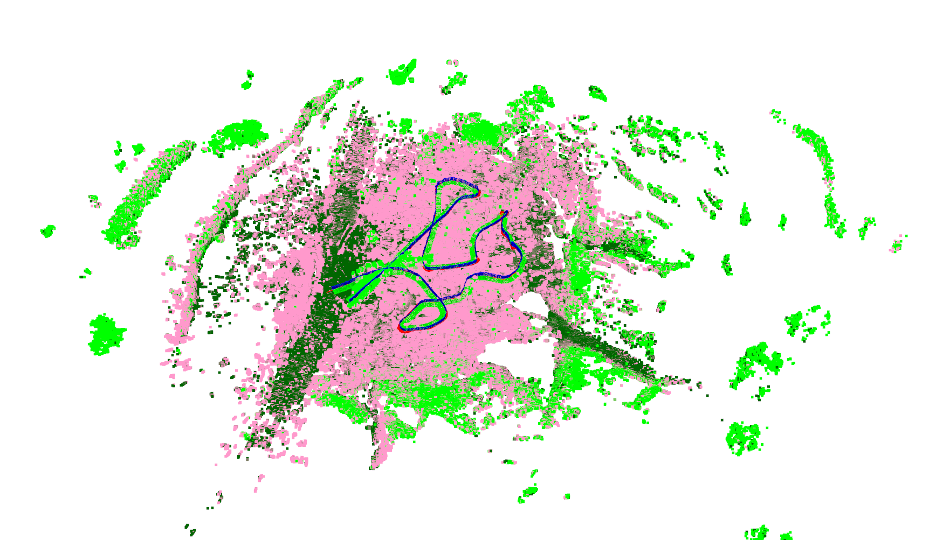}&\includegraphics[width=0.5\textwidth]{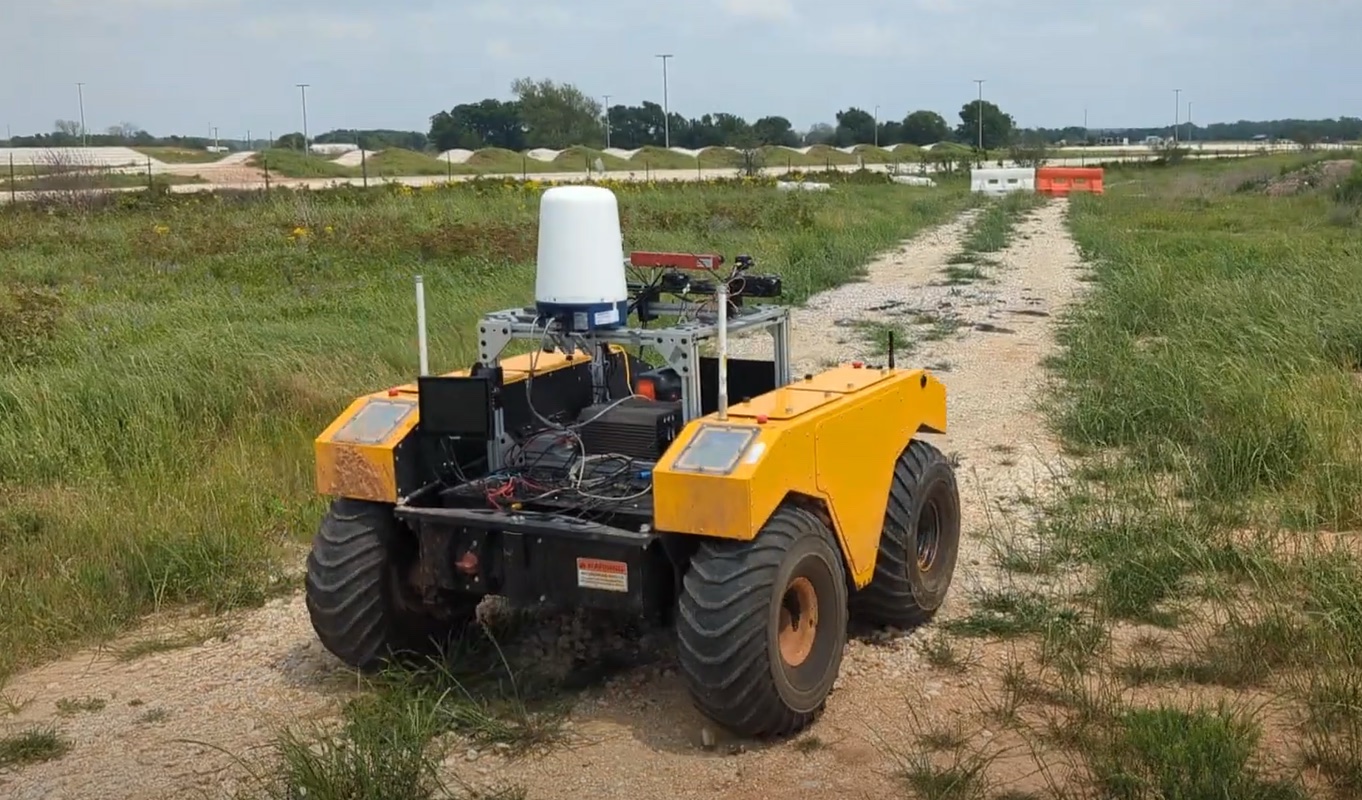}\\
{\scriptsize(a) Point Cloud with Labels} & {\scriptsize(b) Warthog with LIDAR and Radar}\\
\end{tabular} 
\caption{Collected data and experiment platform} 
\label{fig:experiment}
\end{figure}
\section{Technical Approach}
The challenges associated with assigning semantic labels to raw RADAR data, primarily due to its inherent sparsity and the lack of third-dimension information, call for an innovative solution. To tackle this, we propose a comprehensive pipeline that facilitates the transfer of semantic labels from LIDAR to RADAR. This pipeline encompasses several steps as described in Fig. \ref{fig:flow_label} including LIRAR-RADAR calibration, LIDAR-Initial odometry estimation, LIDAR semantic segmentation, 

In terms of addressing the calibration of LIDAR and RADAR, we employ an offline target-based approach, as delineated in \cite{jiang2023improving}. This technique utilizes three aluminum octahedral RADAR reflectors as targets, and employs a Multilayer Perceptron (MLP) to estimate the 3D transformation between the LIDAR and RADAR systems.

To achieve semantic segmentation of the LIDAR data, we deploy the Cylinder3D model \cite{zhouCylinder3DEffective3D2020}. We initially train Cylinder3D on publicly available off-road LIDAR data, after which we perform inference on the collected LIDAR data. We selected Cylinder3D due to its ability to account for the 3D properties of LIDAR, thereby enhancing its generalizability against sensor changes.

To estimate LIDAR odometry, we use the approach described in \cite{xuFASTLIOFastRobust2021,xuFASTLIO2FastDirect2022} to accumulate LIDAR scans. The resultant dense 3D point cloud, derived from the accumulated LIDAR scans, is then voxelized, assigning semantic labels to each voxel grounded in the frequency of labels of points within the voxel. This final point cloud map can be visualized in Fig.\ref{fig:experiment} (a).

\begin{figure}[t]
\centering
\begin{tabular}{c c c} 
\includegraphics[width=0.3\textwidth]{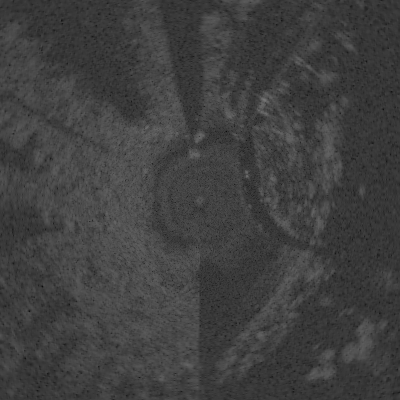}&\includegraphics[width=0.3\textwidth]{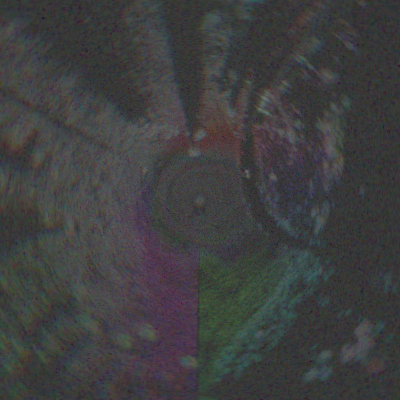}&\includegraphics[width=0.3\textwidth]{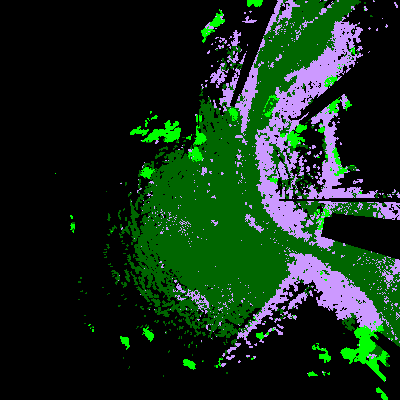}\\
\scriptsize{(a) Single Channel Input}  & \scriptsize{(b) Three Channels Input} & \scriptsize{Generated Label}\\
\end{tabular} 
\caption{RADAR Semantic Confusion Confusion Matrix} 
\label{fig:input_label}
\end{figure}

Subsequently, the LIDAR labels are projected onto the RADAR data using the extrinsic transformation parameters established between LIDAR and RADAR. Upon obtaining the semantic labels of the RADAR data, a semantic segmentation model can be trained, treating the newly generated labels as the ground truth (see Fig. \ref{fig:input_label} (c)). 

The method used to represent RADAR data when training a model can profoundly impact its performance. In semantic segmentation tasks, spatial information is of paramount importance. To preserve the spatial relationship, the RADAR data is projected onto the xy plane of the RADAR coordinate, with the RADAR return energy serving as the pixel value (see Fig.\ref{fig:input_label}(a)).
\subsection{Experiments and Results}
\begin{figure}
\centering
\begin{tabular}{c c} 
\includegraphics[width=0.5\textwidth]{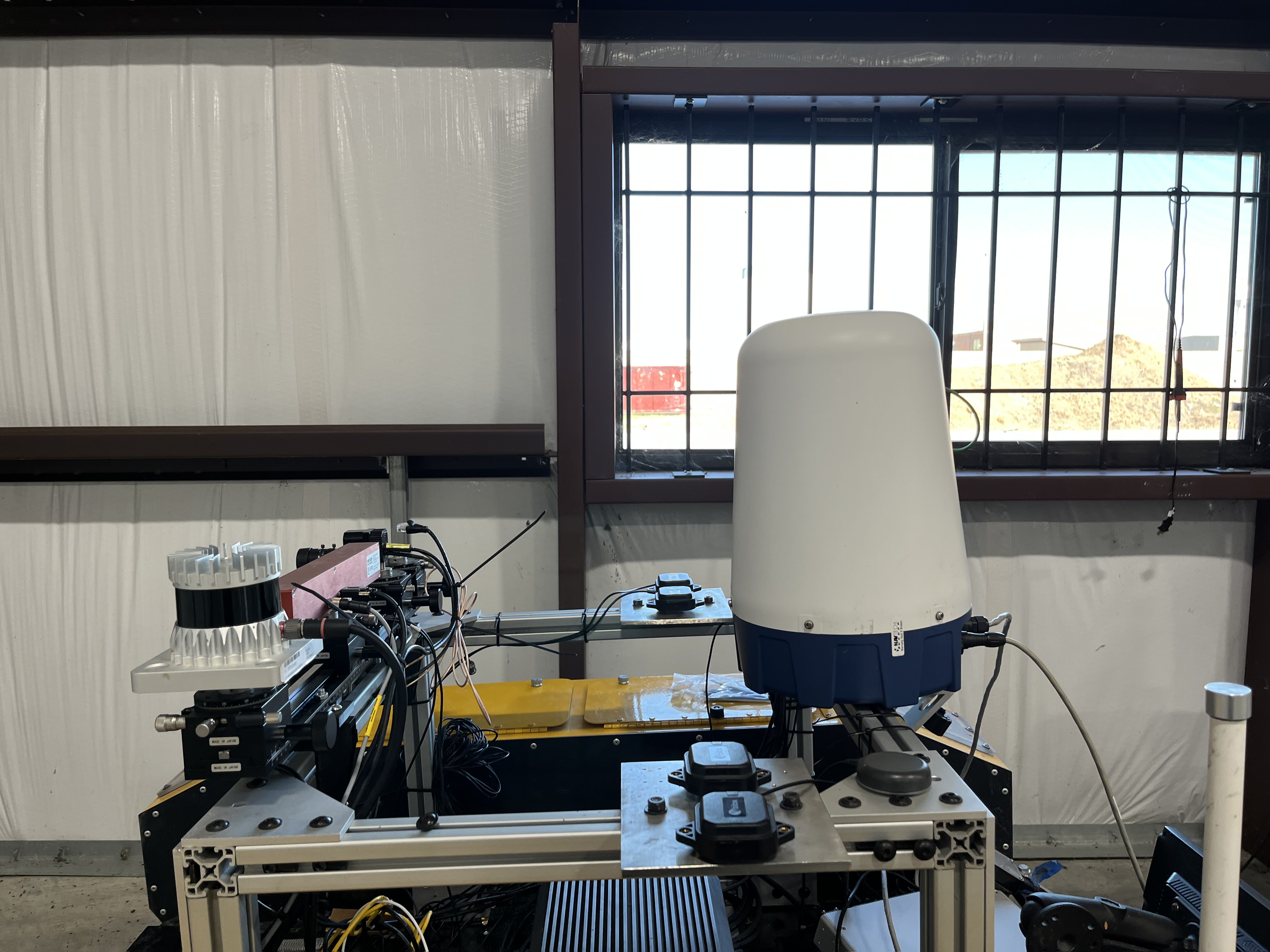}&\includegraphics[width=0.5\textwidth]{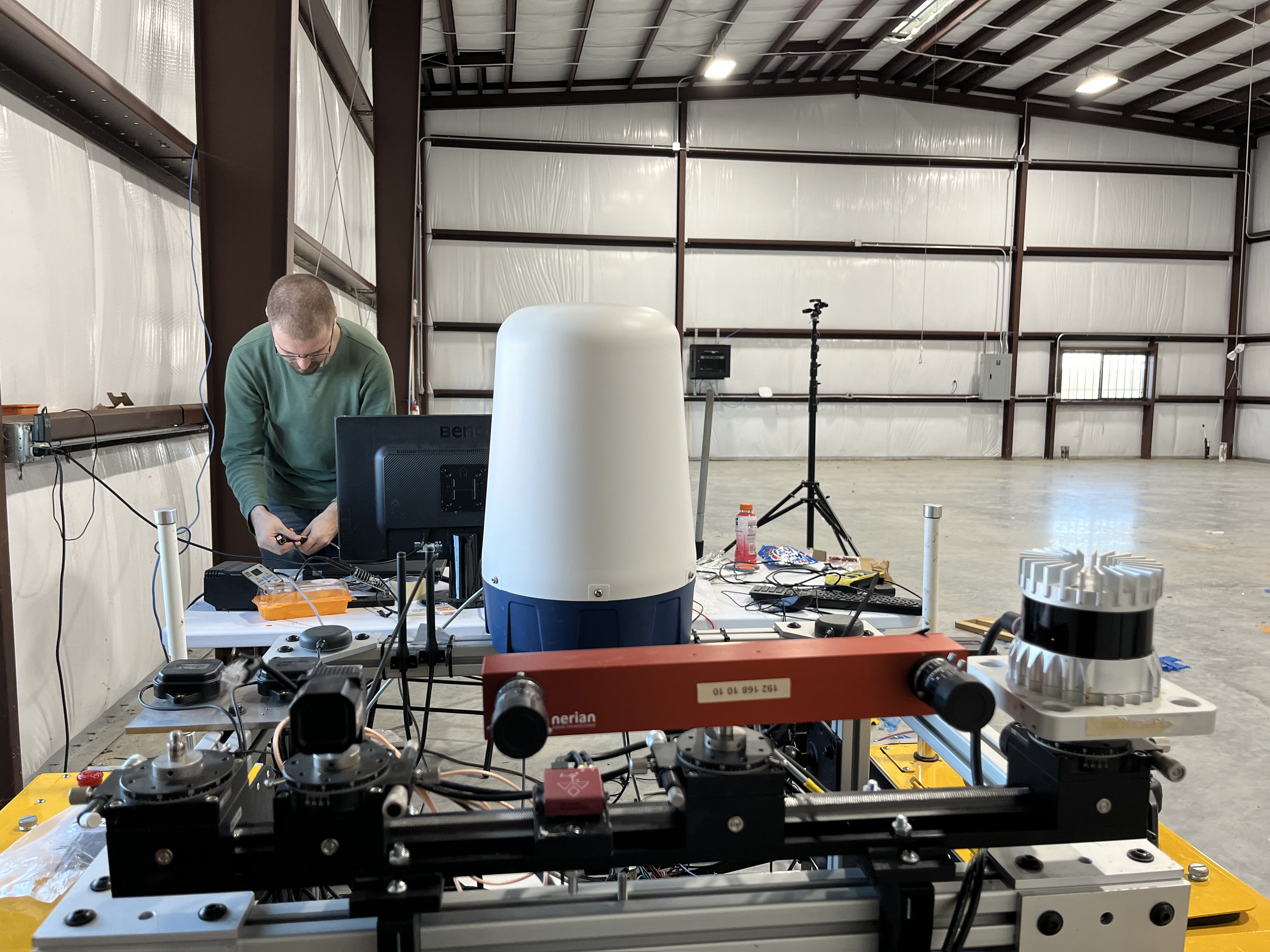}\\
{\scriptsize(a) Side view} & {\scriptsize(b) Front view}\\
\end{tabular} 
\caption{Clearpath Warthog with one Ouster OS1 128 Channel LIDAR and one Navtech CIR-DEV RADAR} 
\label{fig:sensors}
\end{figure}

In order to ascertain the efficacy of our methodology, we have implemented it to test on a real-world dataset. Data collection was carried out in off-road environments, using a Warthog robot equipped with a 128-channel Ouster LIDAR and a Navtech CR-DEV RADAR (see Fig.\ref{fig:sensors}. We collected the data on three different days and totally got 7 sequences. For a comprehensive evaluation, the collected data were divided into two subsets dedicated to training (5 sequences including 5940 RADAR frames) and evaluation (2 sequences 1917 RADAR frames) purposes, respectively.
To facilitate training in LIDAR semantic segmentation, we used the RELLIS-3D dataset\cite{Jiang2021_RELLIS3D} which contains 20 classes. However, it is essential to note that RADAR data differ significantly from LIDAR or image data in terms of sparsity and information loss, which complicates the distinction between various classes. To overcome this challenge, we reduce the number of classes to four, void, bushes, obstacles, and ground, which simplifies the model learning process and improves prediction accuracy. The merged classes are shown as follows:
\begin{itemize}
\item \textbf{Ground}: Grass, Dirt, Asphalt, Concrete, Puddle, Mud
\item \textbf{Bushes}: Bushes
\item \textbf{Obstacles}: Vehicle, Barrier, Log, Pole, Object, Building, Person, Fence, Tree, Rubble
\end{itemize}
\begin{figure}[!t]
\centering
\includegraphics[width=\textwidth]{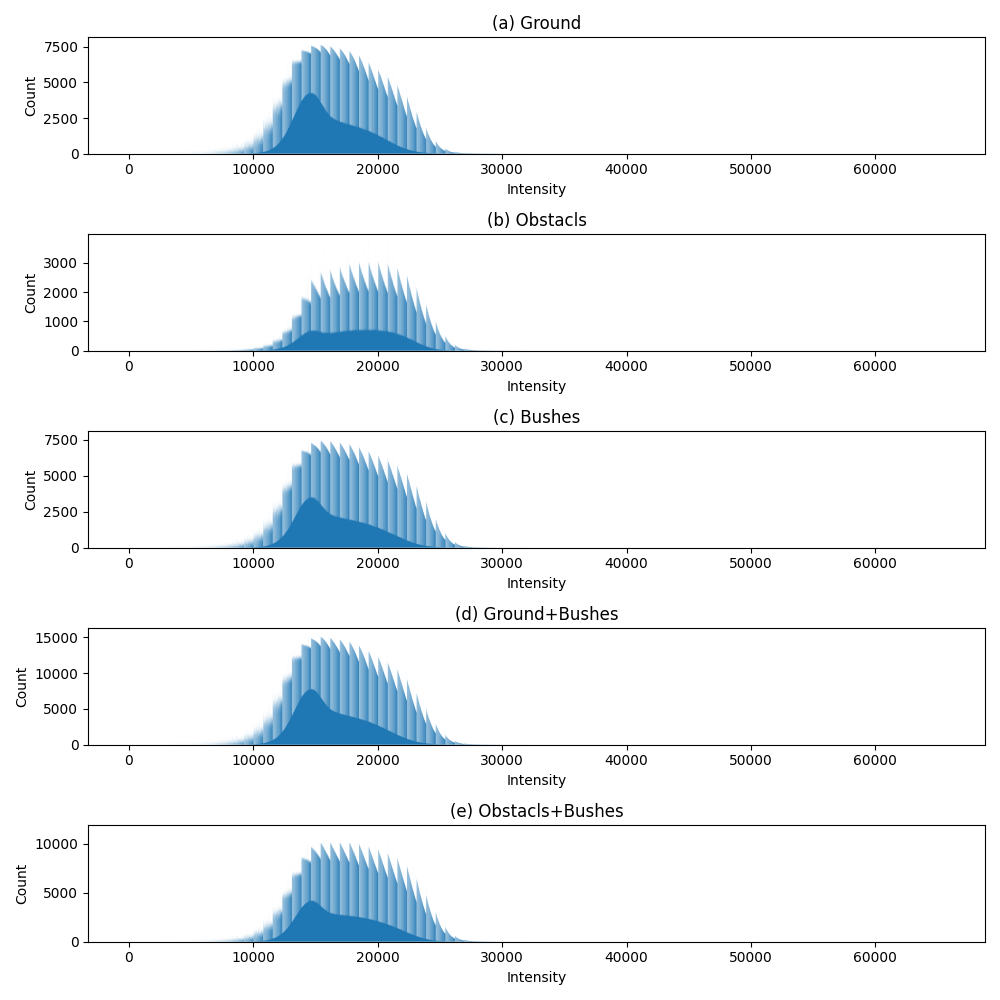}
\caption{Radar intensity distribution for different class aggregations. (a)Ground, (b)obstacles, (c) bushes (d)obstacles and bushes (e)ground and bushes} 
\label{fig:intensity_distribution}
\end{figure}

\subsubsection{Class Intensity Distribution Analysis}\label{sssc:class_intensity_distributin_analysis} Traditionally, RADAR processing has employed techniques such as simple thresholding or the use of a Constant False Alarm Rate (CFAR) filter combined with non-maximal suppression to differentiate between individual returned targets. This inspired us to experiment whether we are able to identify an appropriate threshold that would assist us in discerning the class information.

Figs. \ref{fig:intensity_distribution}(a)-(c) presents the histogram of the RADAR intensity values for three combined classes: Ground, Bushes and Obstacles. While the entire spectrum of RADAR intensity values extends from $0$ to $65535$, it is evident that the majority of these values for all classes are concentrated within the $10000-30000$ range. A notable observation is the overlapping distribution of intensity values for Bushes and Ground. This observation was unexpected; we initially hypothesized that bushes, being a form of obstacle, would exhibit a distribution pattern more aligned with other dense obstacles. On the contrary, the distribution of obstacles tends to be relatively uniform in the range $10000-30000$.

Fig. \ref{fig:intensity_distribution}(d) illustrates the distribution resulting from the amalgamation of Bushes and Obstacles. This combined distribution assumes a bell-shaped curve. Similarly, Figure \ref{fig:intensity_distribution}(e) presents the distribution that emerges after the merging of Ground and Obstacles, which also follows a bell-shaped pattern.

However, these distributions underscore the inherent challenge of pinpointing a singular threshold that can reliably differentiate between the three (or even two) classes. The overlapping and intertwined intensity values suggest that a more sophisticated classification approach might be necessary to achieve satisfactory differentiation.

\subsubsection{Deep learning based classification} 
\begin{figure}[t]
\centering
\begin{tabular}{c c c} 
\includegraphics[width=0.33\textwidth]{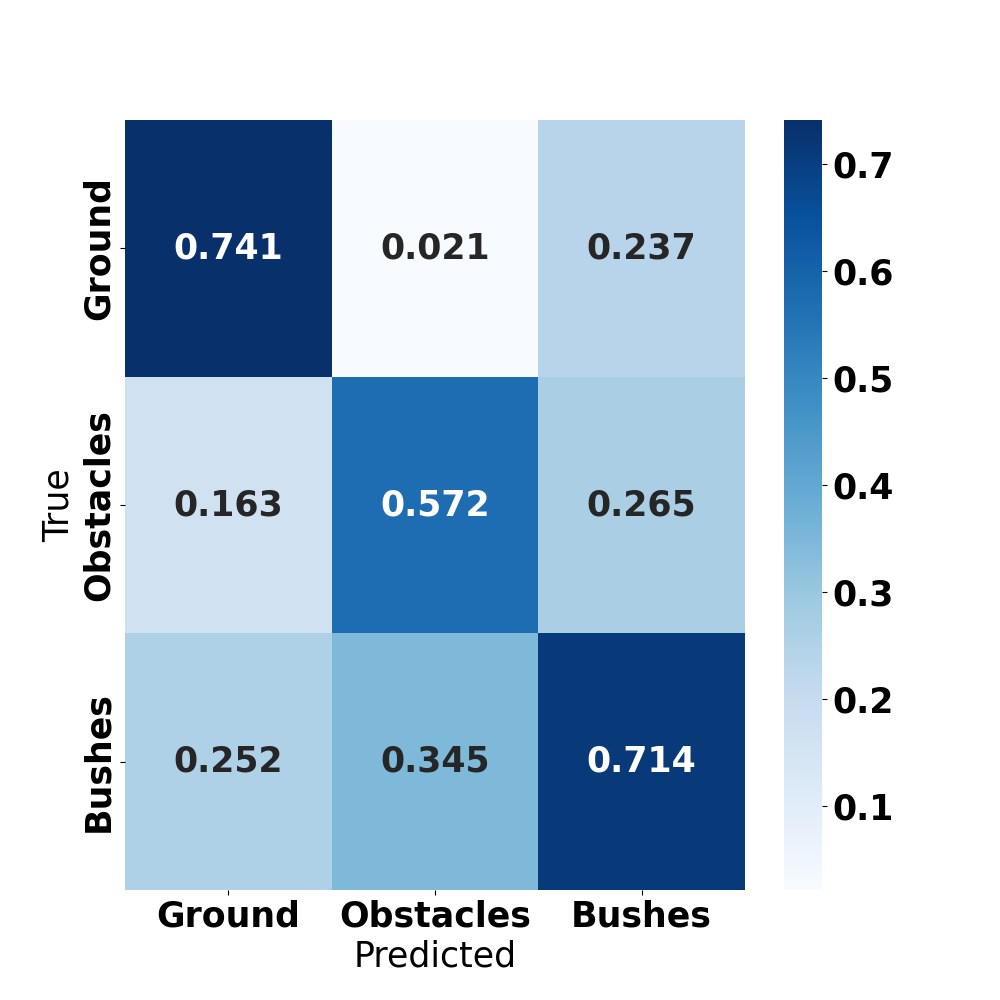}&\includegraphics[width=0.33\textwidth]{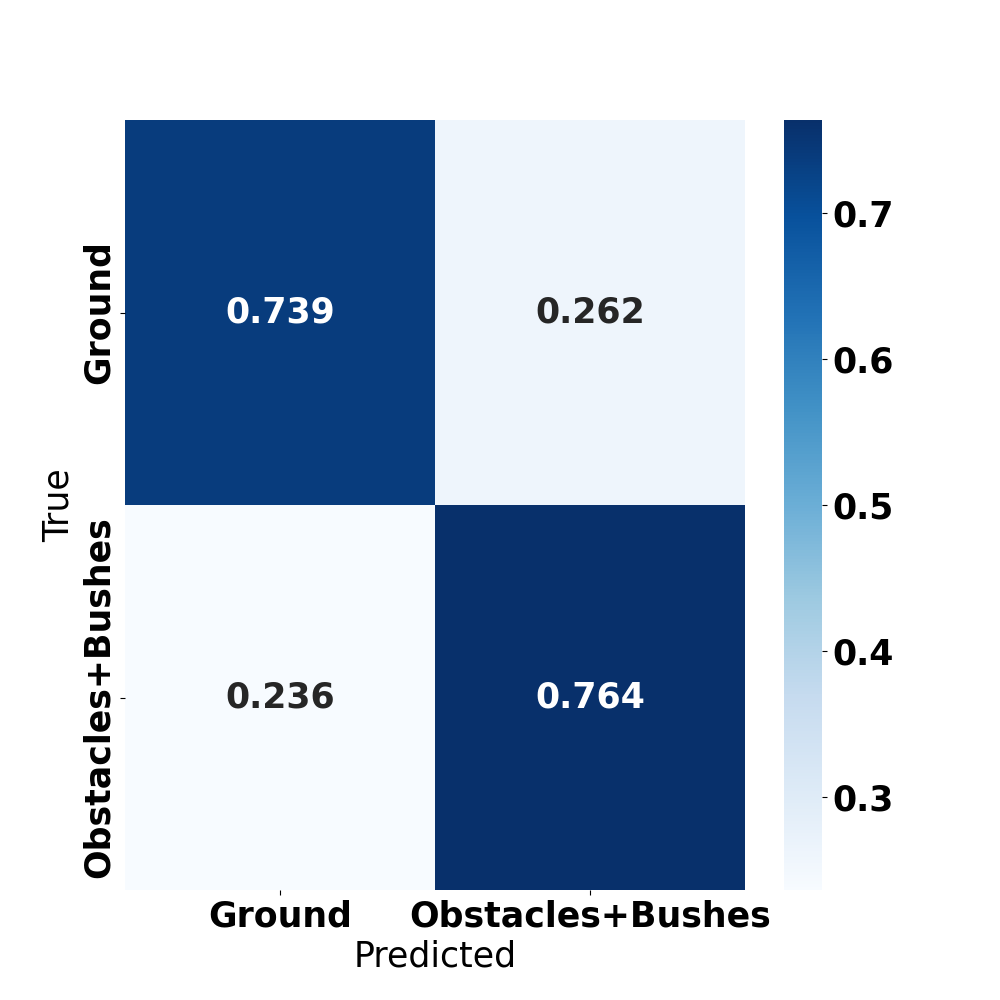}&\includegraphics[width=0.33\textwidth]{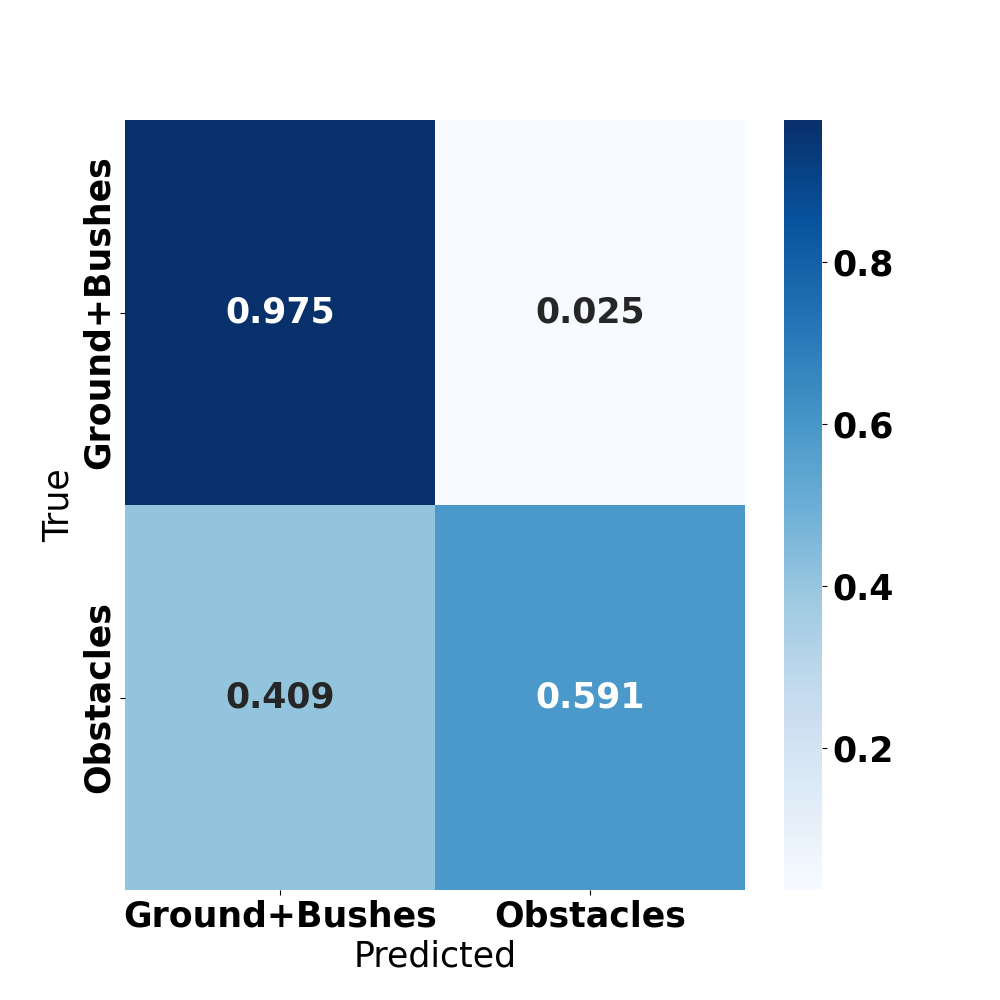}\\
(a) Ch1Cls3 & (b) Ch1Cls2-1 & (c) Ch1Cls2-2\\
\includegraphics[width=0.33\textwidth]{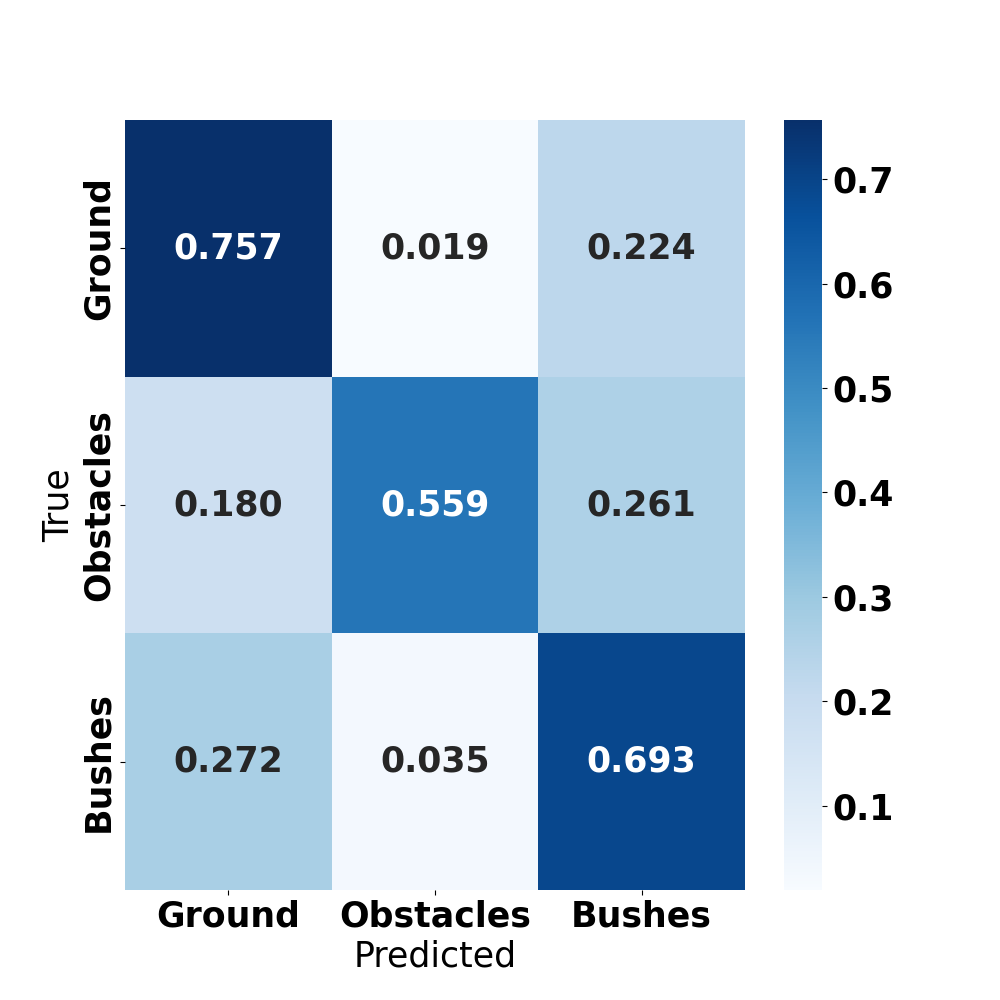}&\includegraphics[width=0.33\textwidth]{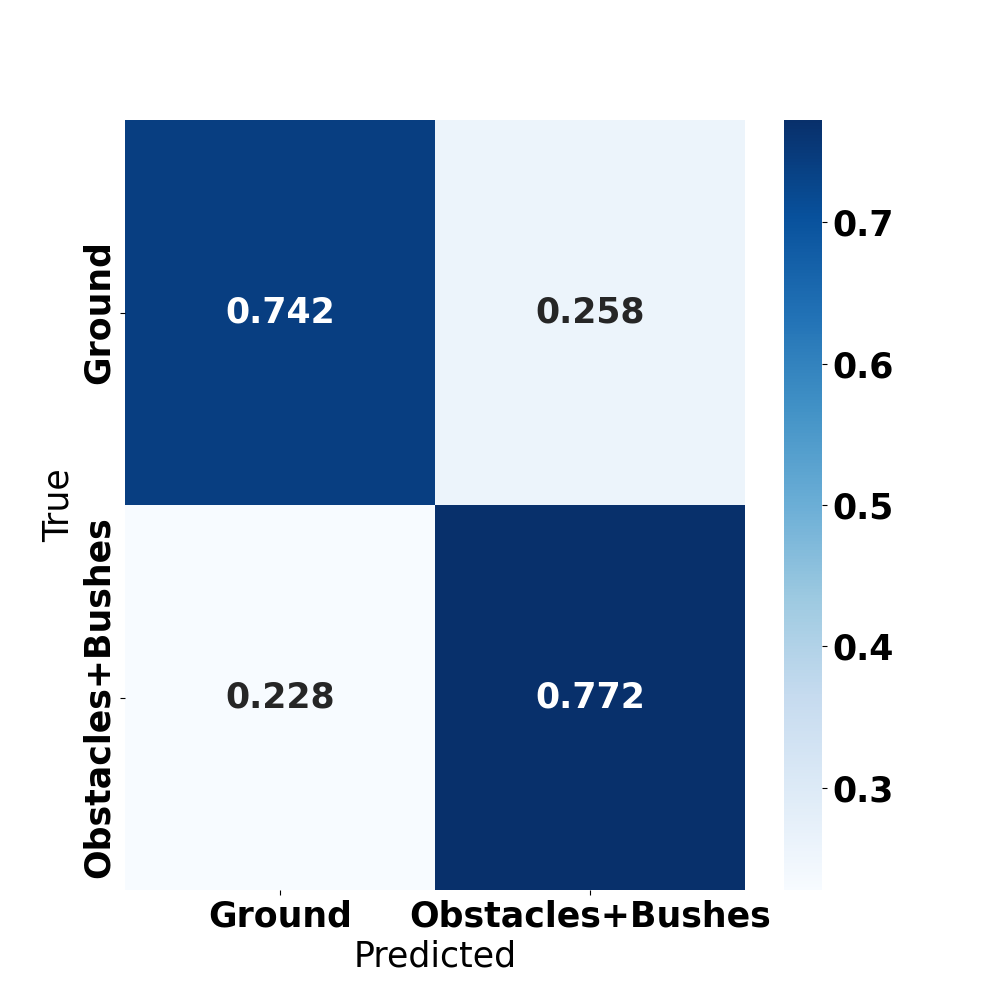}&\includegraphics[width=0.33\textwidth]{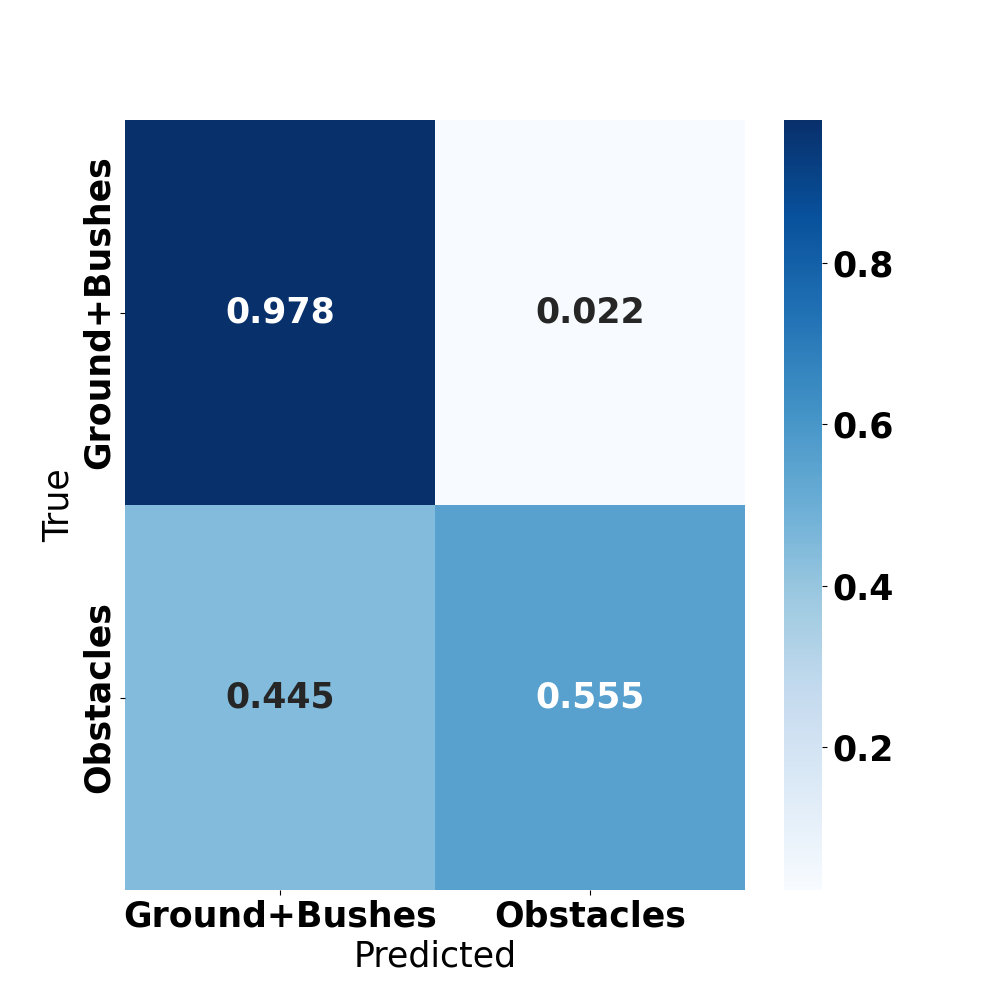}\\
(e) Ch3Cls3 & (f) Ch3Cls2-1 & (g) Ch3Cls2-2\\
\end{tabular} 
\caption{RADAR Semantic Confusion Confusion Matrix} 
\label{fig:radar_confusion_mtx}
\end{figure}

In order to segment the RADAR data, we gravitated toward the acclaimed DeepLabv3 model. Our experiment setup revolved around two types of input: an individual frame (as shown in Fig. \ref{fig:input_label} (a)) and a sequence of three accumulated frames (depicted in Fig. \ref{fig:input_label} (b)), their arrangement based on LIDAR-derived RADAR pose. Observation in \label{sssc:class_intensity_distributin_analysis} guided us towards designing models that segmented RADAR data into two predominant classes: (Ground, Obstacles+Bushes) and (Ground+Bushes, Obstacles). These decisions lead to six distinctive model configurations: Ch1Cls3, Ch3Cls3, Ch1Cls2-1, Ch3Cls2-1, Ch1Cls2-2, and Ch3Cls2-2. The prefix Chx indicates the model's input features x channels, and Clsx defines the output class configurations: Cls3 for (Ground, Obstacles, Bushes), Cls2-1 for (Ground, Obstacles+Bushes), and Cls2-2 for (Ground+Bushes, Obstacles). The experimental results are shown in Table.\ref{tab:miou_table}
 and Fig.\ref{fig:radar_confusion_mtx}
 
Delving into the results, our three-class configuration, Ch1Cls3, revealed notable accuracies: 74. 13\% for Ground, 57. 19\% for Obstacles, and 71. 37\% for Bushes. However, the IoU metrics provided a deeper perspective.  Ch1Cls3 and Ch3Cls3 models showcased similar performances for the ground class, each with an IoU close to 52\%. However, the Ch3Cls3 model, fed with three frames, exhibited a pronounced edge in the Bushes segmentation, boasting an IoU of 53\%, a stark contrast to Ch1Cls3 39. 37\%. It also outperformed the latter in discerning Obstacles, with an IoU of 46.67\% as opposed to Ch1Cls3 41.86\%. These observations were mirrored in the mIoU and mAcc metrics, supporting the argument that leveraging a trio of RADAR frames can refine the segmentation accuracy.

Pivoting to the two-class configurations, the fusion of certain classes unveiled enhanced results. However, the choice of class amalgamation had implications. When blending Ground and Bushes, the Ch1Cls2-2 and Ch3Cls2-2 models achieved stellar accuracies of 97.5\% and 97.8\% respectively for Ground+Bushes. However, the Obstacles class lagged, securing only 59.1\% and 55.5\% accuracy for the two models. In contrast, the fusion of Obstacles and Bushes led to balanced accuracies between the two resultant classes. For example, Ch1Cls2-1 achieved a precision of 73. 9\% for the ground and 76. 4\% for Obstacles + Bushes, while Ch3Cls2-1 recorded 74. 42\% and 77. 2\%, respectively. This parity was mirrored in the mIoU and mAcc metrics. Notably, the introduction of three channels did not consistently enhance the two-class model's performance, underscoring the nuanced role of input depth in segmentation.

\bgroup
\setlength\tabcolsep{0.25cm}
\begin{table}[t]
    \centering
    \begin{tabular}{|c|c|c|c|c|c|} \hline 
         &  Ground(\%)&  Bushes(\%)&  Obstacles(\%)&mIoU(\%)&mAcc(\%)\\ \hline 
         Ch1Cls3&  52.38&  39.37&  41.86&44.54 &61.22\\ \hline 
 Ch3Cls3& 52.11& 53.00& 46.67&50.60 &66.95\\\hline 
 Ch1Cls2-1& 60.55& \multicolumn{2}{c|}{61.94}&61.25&75.98\\ \hline 
 Ch3Cls2-1& 60.43& \multicolumn{2}{c|}{61.39}&60.91&75.71\\ \hline 
         Ch1Cls2-2&  \multicolumn{2}{c|}{69.17}&  57.66&63.42&78.29\\ \hline
 Ch3Cls2-2& \multicolumn{2}{c|}{67.70}& 54.33&61.01&76.67\\ \hline
    \end{tabular}
    \caption{IoU of deep learning-based semantic segmentation}
    \label{tab:miou_table}
\end{table}
\egroup

\section{Experimental Insights}
We conducted experiments on a real-world dataset to gain insights into the connection between semantic segmentation and RADAR data, and to identify potential areas for future research.
\begin{enumerate}

\item  \textbf{Consolidation of Classes}: Our initial experiment into semantic segmentation with RADAR data involved using the full set of classes present in the RELLIS-3D dataset. We quickly observed that our model was struggling to differentiate between these myriad classes. To mitigate this, we consolidated similar terrain features and objects into broader categories (Ground, Bushes, Obstacles), dramatically simplifying the model's task. This approach effectively reduced the granular differentiation that previously overwhelmed the model, which is particularly crucial given the lower spatial resolution of the RADAR data.

\item \textbf{Temporal Component in Input Data}: Our experiments revealed a slight, yet perceptible, enhancement in model performance when we changed from a single frame input to a three-frame sequence input. This implies that temporal data, although they were restricted in our experiment, can have a beneficial effect on model performance. Further research is needed to determine the full extent of this improvement and the most effective way to use temporal information.

\item \textbf{IoU Analysis per Class}: The class-wise IoU results presented interesting trends. The Ground class persistently achieved the highest IoU, indicating that our model excelled at identifying and distinguishing ground surfaces, a promising observation for autonomous navigation tasks. On the contrary, the Obstacle class reported the lowest IoU, indicating a complex and diverse differentiation challenge for the model. This discrepancy may stem from difficulties in identifying the varied objects within the 'Obstacles' class, suggesting the need for a more balanced training dataset or refined model architecture.

\item \textbf{RADAR vs LIDAR}: The experiments underscore the challenges inherent in using RADAR data for semantic segmentation, mainly due to its lower resolution and loss of dimensionality compared to LIDAR. Despite these hurdles, the results are encouraging and hint at the potential for further improvements and adaptations for more robust performance.
\end{enumerate}

\section{Potential Research Avenues for Off-Road Radar Segmentation}
\begin{itemize}

\item \textbf{Redefining Classes for RADAR}: As emphasized in the prior section, the intrinsic resolution and accuracy constraints of RADAR present hurdles in achieving detailed semantic segmentation, particularly when juxtaposed with Cameras and LIDAR. Yet, our experimental evidence accentuates RADAR's potential in broader area segmentation tasks. Given RADAR's distinct attributes, it is prudent to contemplate and introduce RADAR-centric classifications, drawing from aspects like relative height and vegetation density. Additionally, RADAR data provide a lens into the topographical nuances of terrains, discerning features such as slopes and gullies. This insight paves the way for enhanced segmentation avenues, fine-tuned to leverage RADAR's capabilities.

\item \textbf{Alternative Annotation Techniques}: Within the scope of this study, we harnessed both LIDAR and RADAR within a singular robotic system. However, due to LIDAR's restricted range and penetration capabilities, it doesn't encompass the vast expanse detectable by RADAR. For our experimental setup, regions not covered by LIDAR were omitted during both training and evaluation phases, leading to potential data underutilization. Alternative annotation strategies could be explored, such as deploying UAVs for a more expansive and adaptable data collection or tapping into satellite imagery.

\item \textbf{Model Selection and Adaptation}: Our choice for semantic segmentation leaned towards DeepLabv3 \cite{chenRethinkingAtrousConvolution2017}, a model not inherently crafted for RADAR data. However, its commendable performance underscores the adaptability of contemporary segmentation frameworks to diverse sensor data, given the appropriate calibration. For optimal results, the creation of models specifically tuned to RADAR data becomes imperative, ensuring that the features of RADAR are holistically addressed.

\item \textbf{Rethinking RADAR Data Representation and Accumulation}: In our experimental setup, we opted for images as input, capitalizing on the inherently 2D nature of our data. However, our findings underscored the significant contributions of both 3D spatial and temporal information to improving segmentation results. Therefore, exploring alternative RADAR representations that encapsulate richer information, such as voxel maps \cite{timothy2023radar_only}, might prove beneficial.
\end{itemize}

In conclusion, our results highlight the potential of RADAR data in semantic segmentation tasks, especially under challenging off-road environmental conditions. They also point toward significant avenues for further research and improvement to enhance the robustness and applicability of RADAR in the off-road environment.
%
% ---- Bibliography ----
%
\bibliographystyle{splncs03_unsrt}
\bibliography{myReference}

\end{document}